\documentclass[
]{ceurart}

\usepackage{listings}
\usepackage{caption}
\usepackage{subcaption}
\begin{document}

\copyrightyear{2023}
\copyrightclause{Copyright for this paper by its authors.
  Use permitted under Creative Commons License Attribution 4.0   International (CC BY 4.0).}

\conference{CLEF 2023: Conference and Labs of the Evaluation Forum, September 18–21, 2023, Thessaloniki, Greece}

\title{Fraunhofer SIT at CheckThat! 2023: Mixing Single-Modal Classifiers to Estimate the Check-Worthiness of Multi-Modal Tweets}

\title[mode=sub]{Notebook for the CheckThat! Lab at CLEF 2023}


\author[1]{Raphael Antonius Frick}[%
email=raphael.frick@sit.fraunhofer.de,
]

\cormark[1]
\fnmark[1]
\address[1]{Fraunhofer Institute for Secure Information Technology SIT | ATHENE - National Research Center for Applied Cybersecurity,
  Rheinstrasse 75, Darmstadt, 64295, Germany,\\ url=https://www.sit.fraunhofer.de/}

\author[1]{Inna Vogel}[%
email=inna.vogel@sit.fraunhofer.de,
]
\fnmark[1]

\cortext[1]{Corresponding author.}

\begin{abstract}
    The option of sharing images, videos and audio files on social media opens up new possibilities for distinguishing between false information and fake news on the Internet.
    Due to the vast amount of data shared every second on social media, not all data can be verified by a computer or a human expert.
    Here, a check-worthiness analysis can be used as a first step in the fact-checking pipeline and as a filtering mechanism to improve efficiency.
    This paper proposes a novel way of detecting the check-worthiness in multi-modal tweets.
    It takes advantage of two classifiers, each trained on a single modality.
    For image data, extracting the embedded text with an OCR analysis has shown to perform best.
    By combining the two classifiers, the proposed solution was able to place first in the CheckThat! 2023 Task 1A with an $F_1$ score of $0.7297$ achieved on the private test set.
\end{abstract}


\maketitle

\begingroup
\let\clearpage\relax
\section{Introduction}

In today's digitally connected world, social media platforms have become leading channels for the dissemination of information and play a crucial role in shaping public opinion.
However, the proliferation of fake news and false information poses a major challenge to the reliability and trustworthiness of content disseminated on these platforms. 
With billions of data shared on social media platforms every second, it is even for computers infeasible to review all of the data.
There are several AI-based approaches for detecting disinformation on social media platforms.
However, they often do not transfer well to new topics because they have only been trained for specific topics.
Therefore, in many cases, the statements still need to be verified by a human expert.
Aside from using material that has already been fact-checked, a worthiness detection can help to identify content that is worth fact-checking, reducing the computational and human overhead associated with discovering false information.
While much attention has been paid to the detection of review-worthy tweets in text form, detecting content that includes both images and text is still a relatively unexplored area.

Not only is multimedia content frequently shared with text on social media these days, but it can also assist in the spread of disinformation.
They can serve to attract the reader's attention, but also contain false information.
For example, images and videos can be taken out of context and used in a new context, or be manipulated using AI-assisted tools or manual retouching.
In some cases in the past, images consisting only of text were posted without a descriptive text to circumvent the automatic reporting mechanisms of social media platforms such as Facebook. 
This demonstrates the need to extend the check-worthiness estimation from text-only data to multi-modal data.

While the CheckThat! Lab\cite{checkthat_lab} has been trying to solve this problem for text data for several years now, this year a task was offered that revolved around the classification of Twitter posts that contain an image in addition to a text snippet.
Frauhofer SIT participated in Task 1A and 1B of the CLEF 2023 CheckThat! Lab Challenge for the English language. We achieved first place in Task 1A and second place in Task 1B. This paper describes the approach for Task 1A on identifying relevant claims in English multi-modal tweets. 

Our proposed methodology incorporates a multi-modal approach that combines textual cues found in the provided visuals and the descriptive text, leveraging from a pair of  BERT-transformer models to extract meaningful features from them. 

The paper is structured as follows: Section 2 summarises the related work and some winning approaches from the last iterations of the challenge that dealt with textual data. In Section 3, we describe the data consisting multi-modal tweets and that was provided by the CheckThat! Lab organizers. An overview of the tested and proposed approach is given in Section 4 along with their results on the respective data sets. The last section concludes our work with a brief discussion. 
\section{Related Work}

In the past, several approaches have been proposed to estimate the check-worthiness of texts and tweets. 
These methods were based on the extraction of meaningful features. Given a US presidential election transcript, ClaimBuster \cite{Hassan2017} predicts check-worthiness by using an SVM classifier. The authors use the following features: word count, sentiment, tf-idf weighted words, POS tags, and named entities. A total of 6,615 features were used to train the classifier. 

Gencheva~et~al.~\cite{gencheva2017} outperformed the ClaimBuster version by extending the features. Their approach is trained and tested on a data set specially created by the authors based on four US presidential debates from 2016, assigning to every sentence of a document a value between 1 (check-worthy) and 0 (not check-worthy). They aditionally included contextual features such as the position of the sentence, word embeddings, the size of a segment belonging to a speaker or the topic. Using all the features in combination with a neural network (FNN), it outperformed the ClaimBuster version, achieving a MAP of 0.427.   

In the CheckThat! 2018 check-worthiness detection competition, Hansen et al. \cite{hansen2018} showed that a RNN with multiple word representations such as word embeddings, part-of-speech tagging, and syntactic dependencies could achieve state-of-the-art results for check-worthiness detection. The authors later \cite{hansen2019} extended their work and showed significant improvements by applying weak supervision using a collection of unlabelled political speeches. 

The goal of the 2021 CheckThat! challenge was to determine which tweets within a set of COVID-19 related tweets are worth checking. The authors of the best-performing model \cite{martinez2021} fine-tuned several pre-trained transformer models. BERTweet achieved the best results (MAP 0.849 on the development set). Their model was trained on 850 million English tweets and 23 million COVID-19 related English tweets using RoBERTa.  

Savchev \cite{airational} experimented in the 2022 CheckThat! Lab competition with three different pre-trained transformer models: BERT, DistilBERT and RoBERTa. In order to increase the training data, back translation was applied by translating English tweets into French and back into English. The combination of data augmentation and the RoBERTa model gave the best results (F1 0.90, accuracy 0.85) and first place in the competitions. 

Gao et al. \cite{Gao} participated in the AAAI 2022 Multimodal Fact Verification Factify Challenge by implementing two baseline solutions including an ensemble model and an end-to-end multimodal entailment model. The ensemble model outperformed the end-to-end model. They combined two uni-modal models and a multimodal attention network using a 3-way textual entailment classifier, visual similarity with a pre-trained CNN model, and heuristics learned from the dataset. They additionally explored the multimodal fusion technique to model the interaction between different modalities in claim-document pairs and combine information from them. Their best performing model was ranked first by obtaining a weighted average $F_1$ score of $0.77$ on both the validation and test set. 
\section{Data Set Description}

The CheckThat! Lab subtask 1A covered the Arabic and English language; we only participated in the subtask dealing with English data.
The data set consisted of social media posts collected by Twitter through its API. 
Each post featured a text, an image, and an automated OCR analysis provided by Twitter.
Examples from the data set are shown in Figure~\ref{fig:sentExamples}. 

\begin{figure}[h]
	\centering
	\includegraphics[width=0.75\textwidth]{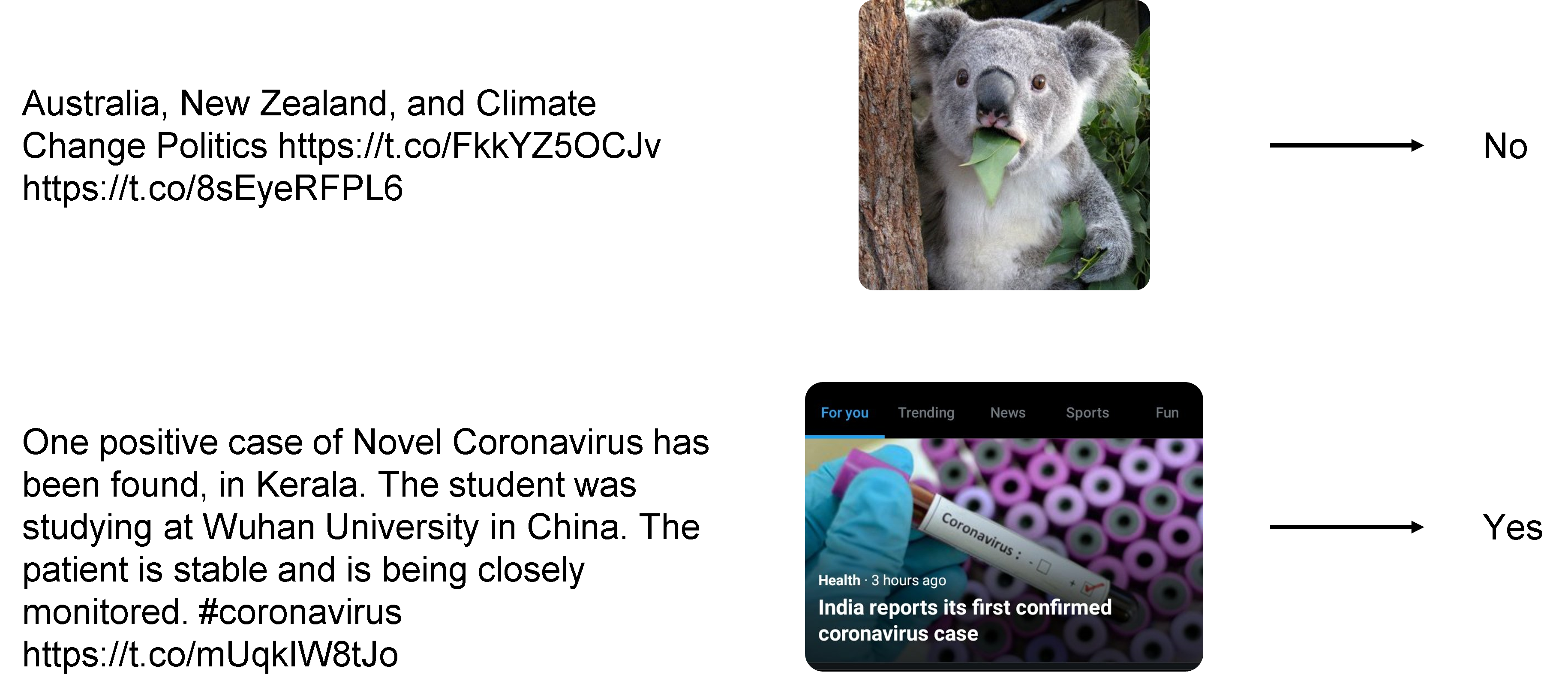}
	
\caption{Instances of check-worthy (Yes) and non-check-worthy (No) tweets for Task 1A} \label{fig:sentExamples}
\end{figure}

The aim of Task 1A was to predict whether a given multi-modal Tweet requires the need of undergoing a manual review by a human expert. 
Along with the contest, a data set was provided that was divided into four splits: a \emph{train} split, a \emph{dev} split, a \emph{dev-test} split, and a \emph{test} split.
While labels for the train set, dev set, and dev-test set were provided upon release, the gold labels for the \emph{test} split were not provided until after the competition was completed.
In addition to the labeled data set, a set of unlabeled data was provided.
The label distributions of each individual data set split are displayed in Table~\ref{tab:class_destribution}. 
As can be seen, the data set suffers from class imbalance.
Within each split, there were almost twice as many tweets not worthy of verification as tweets worthy of verification.

\begin{table}[h]
	\centering
	\caption{Class distribution of the CheckThat! Lab 2023 task 1B English data set}
	\label{tab:class_destribution}
	\begin{tabular}{llll}
		\hline
		& Total  & Yes   & No     \\ \hline
		Train    & 2,356 & 820 & 1,536 \\
		Dev      & 271  & 87 & 184  \\
		Dev Test & 548  & 174   & 374    \\
		Test     &   736   & 277    & 459      \\ \hline
		Sum      & 3,911  & 1,358  & 2,553  \\ \hline
		Unlabeled & 110,173 & ? & ? \\ \hline
	\end{tabular}
\end{table}

\section{Methods and Results}

Detecting check-worthiness in multi-modal tweets presents its own challenges. 
For one, the texts are limited by a character count.
Moreover, both the text and the accompanying image contribute equally to the level of check-worthiness.
In this paper, we present a classification scheme that takes advantage of two classifiers that provide an initial prediction for each modality and then merge their predictions to make a final decision.
By including a step that processes the tweets prior to training and inference, and by using fine-tuning, the classifiers are adapted to the particular writing style typically found in tweets.

\subsection{Pre-Processing}

Unlike text data in documents, books, and web pages, tweets often contain hashtags, emojis, and URLs.
The package \emph{pysentimiento} \cite{perez2021pysentimiento} provides methods for resolving emojis and converting hashtags and URLs to generic tokens.
By resolving emojis into their descriptive meanings, they can be more easily processed by classifiers previously trained on generic text.
Moreover, URLs and hashtags usually do not contribute much to the check-worthiness of a tweet, so their analysis can be omitted.

\begin{table}
\caption{Scores achieved by each model on each data set. $A$ refers to the accuracy score, $P$ to the precision score, $R$ to the recall score and $F_1$ to the $F_1$ score. The character $d$ denotes the dev data split, $dt$ the dev-test data split and $t$ the test split of the data set.}
\label{tab:results}
\begin{tabular}{|c|c|c|c|c|c|}
\hline
\multirow{2}{*}{} & \multirow{2}{*}{\begin{tabular}[c]{@{}l@{}}BERT\end{tabular}} & \multirow{2}{*}{\begin{tabular}[c]{@{}l@{}}BERT \\ + PP\end{tabular}} & \multirow{2}{*}{\begin{tabular}[c]{@{}l@{}}Vision\\Transformer\end{tabular}} & \multirow{2}{*}{OCR} & \multirow{2}{*}{\begin{tabular}[c]{@{}l@{}}BERT + PP\\ + OCR\end{tabular}} \\
                  &                                                                                       &                                                                        &                                                  &                         &                                                                                  \\ \hline

$A_d$     & 0.8155                    &     0.7565       &   0.6790         &    0.7048     &      0.7970                \\ \hline
$P_d$     & 0.7937                    &     0.5827       &   0.0000         &    0.5946     &      0.6379                \\ \hline
$R_d$     & 0.5747                    &     0.8506       &   0.0000         &    0.2529     &      0.8506              \\ \hline
$F1_d$    & 0.6667                    &     0.6916       &   0.0000         &    0.3548     &      0.7291              \\ \hline
$A_{dt}$  & 0.8321                    &    0.7865        &   0.6825          &    0.7190     &      0.8248               \\ \hline
$P_{dt}$  & 0.8361                    &   0.6245         &   0.0000         &    0.6563     &      0.6893                \\ \hline
$R_{dt}$  & 0.5862                    &  0.8218          &   0.0000         &   0.2414     &      0.8161                \\ \hline
$F1_{dt}$ & 0.6892                    &  0.7097          &   0.0000         &   0.3529     &      0.7474                \\ \hline
$A_t$     & 0.7500                    &  0.7772          &   0.6236         &   0.6685     &      0.8057       \\ \hline
$P_t$     & 0.8843                    &  0.6865          &    0.0000        &   0.6701     &      0.7659             \\ \hline
$R_t$     & 0.3863                    &  0.7509          &  0.0000          &   0.2347     &      0.6968         \\ \hline
$F1_t$    & 0.5377                    &  0.7172          &    0.0000        &   0.3476     &      \textbf{0.7297}            \\ \hline
\end{tabular}

\end{table}

\subsection{Classifiying the Textual Data}

To analyze the textual data, a BERT-based\cite{devlin2019bert} model was fine-tuned on the pre-processed tweets. 
Throughout the training process, an optimizer based on the Adam algorithm \cite{kingma2017adam} was employed to take advantage of its adaptive learning rate mechanism. 
Initially, a learning rate of $0.0004$ was selected. The model underwent fine-tuning over five epochs, utilizing a batch size of $24$. 
To ensure optimal performance on the competition data set's development split, only the model checkpoint with the highest performance was retained.

The performance of a BERT model trained with and without pre-processing is displayed in Table \ref{tab:results}.
The classifier trained with pre-processed tweets has higher $F_1$ scores than the one trained without them.
In the specific case of classifying the test set, the $F_1$ score increased from $0.5377$ to $0.7172$.
Thus, it is advisable to take advantage of pre-processing.

\subsection{Classifying the Visual Data}

For the visual data of the data set, a separate classifier was trained.
Two types of classifiers were tried for the challenge, which differed in the type of input data they process: \emph{raw image data} and textual data extracted from an \emph{optical character recognition}.

\paragraph{Using Vision Transformer}

To classify raw image data, a ViT-based Vision Transformer model was fine-tuned \cite{dosovitskiy2021image}.
In particular, the \emph{google/vit-base-patch16-224-in21k} from the huggingface repository was fine-tuned on the provided image train data using a batch size of $16$ within $4$ epochs. 
Similar to the fine-tuned BERT models, Adam was used as the optimizer with a learning rate of $0.0002$ and model checkpoints were utilized.

As shown in Table \ref{tab:results}, the Vision Transformer was unable to learn meaningful patterns.
In particular, the model learned that predicting the majority class ("No") maximizes the validation loss. 
There could be several reasons for this.
Here, the classifier may have tended to classify the majority class due to the class imbalance within the data set, and the images found in each class may not be sufficiently different to provide good class separation.
For further investigation a CNN classifier based on the EfficientNet architecture \cite{tan2020efficientnet} was trained and evaluated.
However, the results did not differ significantly from those of the Vision Transformer.
Therefore, the visual model was not included in the final classifier.

\paragraph{Using Optical Character Recognition}

Since the data-driven imaging models did not perform well in this task, another method for evaluating the information found in the shared images was investigated.

\begin{figure}[h]
	\centering
	\includegraphics[width=0.5\textwidth]{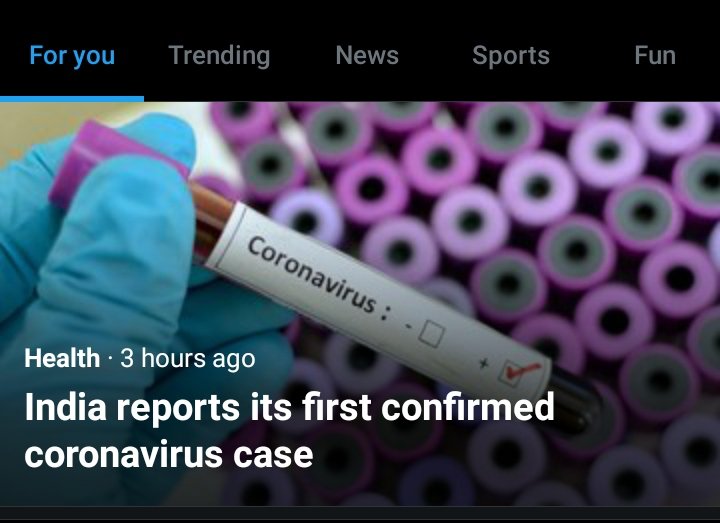}
	
\caption{Text string extracted by easyOCR: \textit{"Trending For you News Sports Fun Coronavirus Health 3 hours ago India reports its first confirmed coronavirus case"}} 
\label{fig:ocr}
\end{figure}

Many times images contain text that can provide additional information for detecting check-worthy content.
To evaluate these, the \emph{easyOCR} package was used.
It is based on the work of Shi et al. \cite{shi2015endtoend} and supports the extraction of text in different languages.
The extracted characters were combined into a single string, which then served as input to a fine-tuned BERT model. 
The BERT model was fine-tuned similar to the classifier predicting the check-worthiness of texts, except a batch size of $8$ was utilized.

Compared to the model that predicts text data, the performance of the classifier that estimates check-worthiness based on text within images provides less good results.
While the accuracy is 69\% on average, the $F_1$ scores obtained for each split are much lower.
Since the data suffers from class imbalance, the $F_1$ score is preferred over the accuracy score.
One reason for the lower scores is that not all images contain text and, on the other hand, some images in the data set were not written in English. 
Therefore, a multilingual model like XLM\cite{lample2019crosslingual} could provide better performance.

\paragraph{Combining BERT with an OCR-Analysis}

For the final solution, the classifier that predicts check-worthiness based on text and the classifier that uses optical character recognition were combined.
First, the validation losses on the dev set were calculated for each classifier.
Then, the loss values were used to weight the predictions made by each individual classification model.
Since the text-based model was able to produce better results, it had a greater impact on the final decision than the image-based model.

Combining the two classifiers resulted in a slight improvement in overall performance.
The $F_1$ values (see Table \ref{tab:results}) were improved across the classification of all data sets.
With a $F_1$ score of $0.7297$ it placed first in the competition.

\include{texfiles/Results} 
\section{Conclusion}

The detection of check-worthy texts can be seen as a first step towards identifying false information spread on the internet.
It can help to reduce the amount of data that needs to be manually reviewed by human experts.
Multi-modal data in social media, such as Twitter, presents new challenges to check-worthiness detection.
The paper presents a new method for detecting review-worthy tweets that contain an image in addition to the descriptive text of the tweet. 
It combines two classifiers trained separately for each modality.
The experiments showed that when analyzing visual data, an OCR analysis outperformed a classifier trained on raw image data. 
Combining the BERT model trained on the tweet text and the BERT model trained on the extracted strings from an optical character recognition slightly improved the performance.
The combined approach performed best in the contest with a $F_1$ score of $0.7297$.
\section*{Acknowledgements}
This work was supported by the German Federal Ministry of Education and Research (BMBF) and the  Hessian Ministry of Higher Education, Research, Science and the Arts within their joint support of “ATHENE – CRISIS” and "Lernlabor Cybersicherheit" (LLCS).
\endgroup

\bibliography{clef23_template}

\begin{thebibliography}{15}
\expandafter\ifx\csname natexlab\endcsname\relax\def\natexlab#1{#1}\fi
\providecommand{\url}[1]{\texttt{#1}}
\providecommand{\href}[2]{#2}
\providecommand{\path}[1]{#1}
\providecommand{\DOIprefix}{doi:}
\providecommand{\ArXivprefix}{arXiv:}
\providecommand{\URLprefix}{URL: }
\providecommand{\Pubmedprefix}{pmid:}
\providecommand{\doi}[1]{\href{http://dx.doi.org/#1}{\path{#1}}}
\providecommand{\Pubmed}[1]{\href{pmid:#1}{\path{#1}}}
\providecommand{\bibinfo}[2]{#2}
\ifx\xfnm\relax \def\xfnm[#1]{\unskip,\space#1}\fi
\bibitem[{Barr{\'o}n-Cede{\~{n}}o et~al.(2023)Barr{\'o}n-Cede{\~{n}}o, Alam,
  Caselli, Da~San~Martino, Elsayed, Galassi, Haouari, Ruggeri, Struss, Nandi,
  Cheema, Azizov, and Nakov}]{checkthat_lab}
\bibinfo{author}{A.~Barr{\'o}n-Cede{\~{n}}o}, \bibinfo{author}{F.~Alam},
  \bibinfo{author}{T.~Caselli}, \bibinfo{author}{G.~Da~San~Martino},
  \bibinfo{author}{T.~Elsayed}, \bibinfo{author}{A.~Galassi},
  \bibinfo{author}{F.~Haouari}, \bibinfo{author}{F.~Ruggeri},
  \bibinfo{author}{J.~M. Struss}, \bibinfo{author}{R.~N. Nandi},
  \bibinfo{author}{G.~S. Cheema}, \bibinfo{author}{D.~Azizov},
  \bibinfo{author}{P.~Nakov},
\newblock \bibinfo{title}{The clef-2023 checkthat! lab: Checkworthiness,
  subjectivity, political bias, factuality, and authority},
\newblock in: \bibinfo{editor}{J.~Kamps}, \bibinfo{editor}{L.~Goeuriot},
  \bibinfo{editor}{F.~Crestani}, \bibinfo{editor}{M.~Maistro},
  \bibinfo{editor}{H.~Joho}, \bibinfo{editor}{B.~Davis},
  \bibinfo{editor}{C.~Gurrin}, \bibinfo{editor}{U.~Kruschwitz},
  \bibinfo{editor}{A.~Caputo} (Eds.), \bibinfo{booktitle}{Advances in
  Information Retrieval}, \bibinfo{publisher}{Springer Nature Switzerland},
  \bibinfo{address}{Cham}, \bibinfo{year}{2023}, pp. \bibinfo{pages}{506--517}.
\bibitem[{Hassan et~al.(2017)Hassan, Arslan, Li, and Tremayne}]{Hassan2017}
\bibinfo{author}{N.~Hassan}, \bibinfo{author}{F.~Arslan},
  \bibinfo{author}{C.~Li}, \bibinfo{author}{M.~Tremayne},
\newblock \bibinfo{title}{Toward automated fact-checking: Detecting
  check-worthy factual claims by claimbuster},
\newblock \bibinfo{journal}{Proceedings of the 23rd ACM SIGKDD International
  Conference on Knowledge Discovery and Data Mining}  (\bibinfo{year}{2017}).
\bibitem[{Gencheva et~al.(2017)Gencheva, Nakov, M{\`a}rquez,
  Barr{\'o}n-Cede{\~n}o, and Koychev}]{gencheva2017}
\bibinfo{author}{P.~Gencheva}, \bibinfo{author}{P.~Nakov},
  \bibinfo{author}{L.~M{\`a}rquez}, \bibinfo{author}{A.~Barr{\'o}n-Cede{\~n}o},
  \bibinfo{author}{I.~Koychev},
\newblock \bibinfo{title}{A context-aware approach for detecting worth-checking
  claims in political debates},
\newblock in: \bibinfo{booktitle}{Proceedings of the International Conference
  Recent Advances in Natural Language Processing, {RANLP} 2017},
  \bibinfo{publisher}{INCOMA Ltd.}, \bibinfo{address}{Varna, Bulgaria},
  \bibinfo{year}{2017}, pp. \bibinfo{pages}{267--276}. \URLprefix
  \url{https://doi.org/10.26615/978-954-452-049-6_037}.
  \DOIprefix\doi{10.26615/978-954-452-049-6_037}.
\bibitem[{Hansen et~al.(2018)Hansen, Hansen, Simonsen, and Lioma}]{hansen2018}
\bibinfo{author}{C.~Hansen}, \bibinfo{author}{C.~Hansen},
  \bibinfo{author}{J.~Simonsen}, \bibinfo{author}{C.~Lioma},
\newblock \bibinfo{title}{The copenhagen team participation in the
  check-worthiness task of the competition of automatic identification and
  verification of claims in political debates of the clef-2018 checkthat! lab},
\newblock in: \bibinfo{editor}{L.~{Cappellato }}, \bibinfo{editor}{N.~{Ferro
  }}, \bibinfo{editor}{J.~Nie}, \bibinfo{editor}{L.~Soulier} (Eds.),
  \bibinfo{booktitle}{CLEF 2018 Working Notes}, CEUR Workshop Proceedings,
  \bibinfo{publisher}{CEUR-WS.org}, \bibinfo{year}{2018}. \bibinfo{note}{19th
  Working Notes of CLEF Conference and Labs of the Evaluation Forum, CLEF 2018
  ; Conference date: 10-09-2018 Through 14-09-2018}.
\bibitem[{Hansen et~al.(2019)Hansen, Hansen, Simonsen, and Lioma}]{hansen2019}
\bibinfo{author}{C.~Hansen}, \bibinfo{author}{C.~Hansen},
  \bibinfo{author}{J.~G. Simonsen}, \bibinfo{author}{C.~Lioma},
\newblock \bibinfo{title}{Neural weakly supervised fact check-worthiness
  detection with contrastive sampling-based ranking loss},
\newblock in: \bibinfo{editor}{L.~Cappellato}, \bibinfo{editor}{N.~Ferro},
  \bibinfo{editor}{D.~E. Losada}, \bibinfo{editor}{H.~M{\"{u}}ller} (Eds.),
  \bibinfo{booktitle}{Working Notes of {CLEF} 2019 - Conference and Labs of the
  Evaluation Forum, Lugano, Switzerland, September 9-12, 2019}, volume
  \bibinfo{volume}{2380} of \textit{\bibinfo{series}{{CEUR} Workshop
  Proceedings}}, \bibinfo{publisher}{CEUR-WS.org}, \bibinfo{year}{2019}.
  \URLprefix \url{https://ceur-ws.org/Vol-2380/paper\_56.pdf}.
\bibitem[{Martinez{-}Rico et~al.(2021)Martinez{-}Rico, Mart{\'{\i}}nez{-}Romo,
  and Araujo}]{martinez2021}
\bibinfo{author}{J.~R. Martinez{-}Rico},
  \bibinfo{author}{J.~Mart{\'{\i}}nez{-}Romo}, \bibinfo{author}{L.~Araujo},
\newblock \bibinfo{title}{Nlp{\&}ir@uned at checkthat!~2021: Check-worthiness
  estimation and fake news detection using transformer models},
\newblock in: \bibinfo{editor}{G.~Faggioli}, \bibinfo{editor}{N.~Ferro},
  \bibinfo{editor}{A.~Joly}, \bibinfo{editor}{M.~Maistro},
  \bibinfo{editor}{F.~Piroi} (Eds.), \bibinfo{booktitle}{Proceedings of the
  Working Notes of {CLEF} 2021 - Conference and Labs of the Evaluation Forum,
  Bucharest, Romania, September 21st - to - 24th, 2021}, volume
  \bibinfo{volume}{2936} of \textit{\bibinfo{series}{{CEUR} Workshop
  Proceedings}}, \bibinfo{publisher}{CEUR-WS.org}, \bibinfo{year}{2021}, pp.
  \bibinfo{pages}{545--557}. \URLprefix
  \url{https://ceur-ws.org/Vol-2936/paper-44.pdf}.
\bibitem[{Savchev(2022)}]{airational}
\bibinfo{author}{A.~Savchev},
\newblock \bibinfo{title}{{AI} rational at checkthat!-2022: Using transformer
  models for tweet classification},
\newblock in: \bibinfo{editor}{G.~Faggioli}, \bibinfo{editor}{N.~Ferro},
  \bibinfo{editor}{A.~Hanbury}, \bibinfo{editor}{M.~Potthast} (Eds.),
  \bibinfo{booktitle}{Proceedings of the Working Notes of {CLEF} 2022 -
  Conference and Labs of the Evaluation Forum, Bologna, Italy, September 5th -
  to - 8th, 2022}, volume \bibinfo{volume}{3180} of
  \textit{\bibinfo{series}{{CEUR} Workshop Proceedings}},
  \bibinfo{publisher}{CEUR-WS.org}, \bibinfo{year}{2022}, pp.
  \bibinfo{pages}{656--659}. \URLprefix
  \url{https://ceur-ws.org/Vol-3180/paper-52.pdf}.
\bibitem[{Gao et~al.(2022)Gao, Hoffmann, Oikonomou, Kiskovski, and
  Bandhakavi}]{Gao}
\bibinfo{author}{J.~Gao}, \bibinfo{author}{H.~Hoffmann},
  \bibinfo{author}{S.~Oikonomou}, \bibinfo{author}{D.~Kiskovski},
  \bibinfo{author}{A.~Bandhakavi},
\newblock \bibinfo{title}{Logically at factify 2022: Multimodal fact
  verfication},
\newblock in: \bibinfo{editor}{A.~Das}, \bibinfo{editor}{T.~Chakraborty},
  \bibinfo{editor}{A.~Ekbal}, \bibinfo{editor}{A.~P. Sheth} (Eds.),
  \bibinfo{booktitle}{Proceedings of the Workshop on Multi-Modal Fake News and
  Hate-Speech Detection {(DE-FACTIFY} 2022) co-located with the Thirty-Sixth
  {AAAI} Conference on Artificial Intelligence {(} {AAAI} 2022), Virtual Event,
  Vancouver, Canada, February 27, 2022}, volume \bibinfo{volume}{3199} of
  \textit{\bibinfo{series}{{CEUR} Workshop Proceedings}},
  \bibinfo{publisher}{CEUR-WS.org}, \bibinfo{year}{2022}. \URLprefix
  \url{https://ceur-ws.org/Vol-3199/paper6.pdf}.
\bibitem[{Pérez et~al.(2021)Pérez, Giudici, and
  Luque}]{perez2021pysentimiento}
\bibinfo{author}{J.~M. Pérez}, \bibinfo{author}{J.~C. Giudici},
  \bibinfo{author}{F.~Luque}, \bibinfo{title}{pysentimiento: A python toolkit
  for sentiment analysis and socialnlp tasks}, \bibinfo{year}{2021}.
  \href{http://arxiv.org/abs/2106.09462}{{\tt arXiv:2106.09462}}.
\bibitem[{Devlin et~al.(2019)Devlin, Chang, Lee, and
  Toutanova}]{devlin2019bert}
\bibinfo{author}{J.~Devlin}, \bibinfo{author}{M.-W. Chang},
  \bibinfo{author}{K.~Lee}, \bibinfo{author}{K.~Toutanova},
  \bibinfo{title}{Bert: Pre-training of deep bidirectional transformers for
  language understanding}, \bibinfo{year}{2019}.
  \href{http://arxiv.org/abs/1810.04805}{{\tt arXiv:1810.04805}}.
\bibitem[{Kingma and Ba(2017)}]{kingma2017adam}
\bibinfo{author}{D.~P. Kingma}, \bibinfo{author}{J.~Ba}, \bibinfo{title}{Adam:
  A method for stochastic optimization}, \bibinfo{year}{2017}.
  \href{http://arxiv.org/abs/1412.6980}{{\tt arXiv:1412.6980}}.
\bibitem[{Dosovitskiy et~al.(2021)Dosovitskiy, Beyer, Kolesnikov, Weissenborn,
  Zhai, Unterthiner, Dehghani, Minderer, Heigold, Gelly, Uszkoreit, and
  Houlsby}]{dosovitskiy2021image}
\bibinfo{author}{A.~Dosovitskiy}, \bibinfo{author}{L.~Beyer},
  \bibinfo{author}{A.~Kolesnikov}, \bibinfo{author}{D.~Weissenborn},
  \bibinfo{author}{X.~Zhai}, \bibinfo{author}{T.~Unterthiner},
  \bibinfo{author}{M.~Dehghani}, \bibinfo{author}{M.~Minderer},
  \bibinfo{author}{G.~Heigold}, \bibinfo{author}{S.~Gelly},
  \bibinfo{author}{J.~Uszkoreit}, \bibinfo{author}{N.~Houlsby},
  \bibinfo{title}{An image is worth 16x16 words: Transformers for image
  recognition at scale}, \bibinfo{year}{2021}.
  \href{http://arxiv.org/abs/2010.11929}{{\tt arXiv:2010.11929}}.
\bibitem[{Tan and Le(2020)}]{tan2020efficientnet}
\bibinfo{author}{M.~Tan}, \bibinfo{author}{Q.~V. Le},
  \bibinfo{title}{Efficientnet: Rethinking model scaling for convolutional
  neural networks}, \bibinfo{year}{2020}.
  \href{http://arxiv.org/abs/1905.11946}{{\tt arXiv:1905.11946}}.
\bibitem[{Shi et~al.(2015)Shi, Bai, and Yao}]{shi2015endtoend}
\bibinfo{author}{B.~Shi}, \bibinfo{author}{X.~Bai}, \bibinfo{author}{C.~Yao},
  \bibinfo{title}{An end-to-end trainable neural network for image-based
  sequence recognition and its application to scene text recognition},
  \bibinfo{year}{2015}. \href{http://arxiv.org/abs/1507.05717}{{\tt
  arXiv:1507.05717}}.
\bibitem[{Lample and Conneau(2019)}]{lample2019crosslingual}
\bibinfo{author}{G.~Lample}, \bibinfo{author}{A.~Conneau},
  \bibinfo{title}{Cross-lingual language model pretraining},
  \bibinfo{year}{2019}. \href{http://arxiv.org/abs/1901.07291}{{\tt
  arXiv:1901.07291}}.

\end{thebibliography}

\appendix

\end{document}